\documentclass[twocolumn,10pt]{jmlr} 

\usepackage{booktabs}

\usepackage{siunitx}
\usepackage{multirow}
\usepackage{bbm}
\usepackage{svg}
\usepackage{arydshln}
\newcommand{\red}[1]{\textcolor{red}{#1}}
\newcommand{\green}[1]{\textcolor{blue}{#1}}

\title[Neural Fine-Gray]{Neural Fine-Gray:\\ Monotonic neural networks for competing risks}

\author{
    \Name{Vincent Jeanselme} \Email{vincent.jeanselme@mrc-bsu.cam.ac.uk}\\
    \addr The Alan Turing Institute\\
    \addr MRC Biostatistics Unit, University of Cambridge
\AND
    \Name{Chang Ho Yoon} \Email{changho.yoon@sjc.ox.ac.uk}\\
    \addr The Alan Turing Institute\\
    \addr Big Data Institute, University of Oxford
\AND
    \Name{Brian Tom} \Email{brian.tom@mrc-bsu.cam.ac.uk}\\
    \Name{Jessica Barrett} \Email{jessica.barrett@mrc-bsu.cam.ac.uk}\\
    \addr MRC Biostatistics Unit, University of Cambridge
}

\usepackage{pbox}

\begin{document}

\maketitle


\begin{abstract}
Time-to-event modelling, known as survival analysis, differs from standard regression as it addresses \emph{censoring} in patients who do not experience the event of interest. Despite competitive performances in tackling this problem, machine learning methods often ignore other \emph{competing risks} that preclude the event of interest. This practice biases the survival estimation. Extensions to address this challenge often rely on parametric assumptions or numerical estimations leading to sub-optimal survival approximations. This paper leverages constrained monotonic neural networks to model each competing survival distribution. This modelling choice ensures the exact likelihood maximisation at a reduced computational cost by using automatic differentiation. The effectiveness of the solution is demonstrated on one synthetic and three medical datasets. Finally, we discuss the implications of considering competing risks when developing risk scores for medical practice.
\end{abstract}

\paragraph*{Data and Code Availability}
Experiments are performed on publicly available datasets: Primary Biliary Cholangitis\footnote{Available in the R survival package.}~\citep{therneau2000cox}, Framingham\footnote{Available in the R riskCommunicator package.}~\citep{kannel1979diabetes}, Synthetic\footnote{Available at \url{https://github.com/chl8856/DeepHit}}~\citep{lee2018deephit}, and the Surveillance, Epidemiology, and End Results Program\footnote{Available at \url{https://seer.cancer.gov/}}. The code to reproduce the proposed model and the presented results is available on GitHub\footnote{\url{https://github.com/Jeanselme/NeuralFineGray}}.

\paragraph*{Institutional Review Board (IRB)}
This research does not require IRB approval as it relies on publicly available datasets from previous studies.


\section{Introduction}

\subsection{Motivation}
Survival analysis involves modelling the time to an event of interest, which plays a critical role in medicine to understand disease manifestation, treatment outcomes, and the influence of different risk factors on patient health~\citep{selvin2008survival}. This analysis differs from standard regression settings as patients may not experience the outcome of interest over the study period. These \textit{censored} patients inform this regression as they participate in the study event-free until exiting the study. Multiple approaches have been proposed to take advantage of these patients by maximising the likelihood of the observed data.

Often, in medical data, patients may experience events, known as \emph{competing risks}, that preclude the observation of the event of interest. For instance, in modelling the time to cardiac events, patients who die from another condition during the observation period exit the study because of a competing risk. Competing risks remain overlooked despite their prevalence in medicine~\citep{koller2012competing, austin2016introduction}. Particularly, practitioners frequently consider competing risks as censoring~\citep{austin2017accounting}. This practice breaks the common assumption of non-informative censoring,~i.e.,~censored patients must leave the study for reasons independent of the outcome of interest. Considering competing risks as censoring, therefore, results in misestimating the risk of the event of interest~\citep{fisher1974presenting, leung1997censoring}.

To better tackle the problem of competing risks, one can explicitly model them through the marginal probability of observing each risk, known as the Cumulative Incidence Function (CIF). Estimation of these functions often relies on proportional hazards, parametric assumptions, or numerical integration, potentially resulting in the optimisation of a sub-optimal target misrepresenting the true underlying survival distribution. 

\subsection{Contribution}
This work introduces a novel machine learning model to tackle the problem of competing risks. This approach generalises~\cite{rindt2022survival} to competing risks, leveraging monotonic neural networks to model cumulative incidence functions. The proposed method tackles the limitations of existing strategies by an exact computation of the likelihood at a lower computational cost.

First, we explore the existing literature before introducing in detail our proposed model. Subsequently, we demonstrate the advantages and limitations of our approach as applied to one synthetic and three real-world medical datasets. Finally, we further investigate the Framingham dataset to underline the importance of considering competing risks in cardiovascular disease risk estimation.

\section{Related work}
This section summarises the recent progress in machine learning for survival analysis.

\subsection{Time-to-event modelling}
Survival analysis is an active field of research in the statistical community~\citep{kartsonaki2016survival}. Non-parametric~\citep{ishwaran2008random} and parametric~\citep{cox2008generalized, royston2001flexible, cox2007parametric} models have been introduced to model survival outcomes. Despite these multiple alternatives and considerable proposed extensions, the original Cox proportional-hazards model~\citep{cox1972regression} remains widely used in the medical literature~\citep{stensrud2020test}. This semi-parametric approach estimates the impact of covariates on the instantaneous risk of observing an event,~i.e.,~hazard. The model assumes the hazard to take the form of the product of a non-parametric estimate of the population survival and a parametric covariate effect. This assumption is known as proportional hazards and renders tractable the model optimisation for covariate effect estimation.

The machine learning community has extended the Cox model for unknown parametric forms of covariate effect. Specifically, DeepSurv~\citep{katzman2018deepsurv} replaces this otherwise parametric component with a neural network. However, this model still assumes proportional hazards that may not hold in real-world medical settings~\citep{stensrud2020test}. To relax this assumption, DeepCox~\citep{nagpal2021deep} identifies subgroups using independent Cox models. Each subgroup is characterised by its own non-parametric baseline and covariate effect. At the intersection between DeepCox and parametric models,~\cite{nagpal2020deep} model each subgroup with a Weibull distribution parameterised by neural networks to allow end-to-end training. \cite{jeanselme2022neural} abandon the parametric and proportional hazards assumption with unconstrained distributions learnt through monotonic networks.

With a focus on predictive performance, DeepHit~\citep{lee2018deephit} approaches survival as a classification problem where survival prediction time is discretised. The associated task is to predict the interval at which a patient experiences the event. The model's training procedure consists of a likelihood and a ranking penalty which favours temporally coherent predictions. Extrapolation of this model to infinite time discretisation resembles an ordinary differential equation (ODE), as proposed in~\cite{danks2022derivative}.

The models above approximate the underlying survival likelihood either through parametric assumptions, discretisation or numerical integration. Recently,~\cite{rindt2022survival} proposed to overcome this challenge of likelihood estimation by deploying a constrained neural network with a monotonically increasing outcome to obtain the survival function, and, therefore, the exact likelihood. In addition, to show improved performance, the authors demonstrate that one should prefer likelihood optimisation over discriminative performance as the optimal likelihood is obtained for the true underlying survival distribution,~i.e.,~the likelihood is a proper scoring rule. Our study is a generalisation of this work to competing risks, harnessing monotonic neural networks to directly model CIFs.

\subsection{Modelling competing risks}
Using the aforementioned models without consideration of competing risks would lead to a misestimation of the risk associated with the event of interest~\citep{schuster2020ignoring}. To tackle this issue, one can independently estimate each competing-risk-specific model and combine them to estimate the risk associated with a specific outcome given the non-observation of the other risks, as formulated in the cause-specific Cox model~\citep{prentice1978analysis}. This independent estimation describes how covariates impact each event risk~\citep{austin2017practical} but may misrepresent the relative effect of these covariates on outcomes~\citep{austin2016introduction} and lead to sub-optimal predictive performance. Alternatively, \cite{fine1999proportional} propose to model the sub-hazards, i.e.,~the probability of observing a given event if the patient has not experienced this event until $t$, under an assumption of proportionality analogous to the one made in the Cox proportional-hazards model. While providing insights into the link between covariates and risk particularly suitable for prediction~\citep{austin2017practical}, this model suffers from two shortcomings: (i)~the proportionality assumption impairs its real-world applicability; (ii)~this approach can result in an ill-defined survival function~\citep{austin2021fine}.

Machine learning approaches have been extended to jointly model competing risks. DeepHit's time-discretisation results in a straightforward extension in which the output dimension is multiplied by the number of risks~\citep{lee2018deephit}. Similarly, hierarchical discretisation~\citep{tjandra2021hierarchical} has been proposed. As parametric distributional assumptions result in a closed-form likelihood, \cite{nagpal2020deep} propose to extend their mixture of Weibull distributions and~\cite{bellot2018tree} introduce a Bayesian mixture of Generalised Gamma distributions to tackle competing risk. Under more complex non-parametric likelihoods, numerical integration~\citep{danks2022derivative, aastha2020deepcompete} and pseudo-value approximations~\citep{rahman2021deeppseudo} have been proposed. Finally, non-likelihood-based approaches have been introduced such as boosted trees~\citep{bellot2018multitask} or survival trees~\citep{schmid2021competing}. However, these methods are optimised towards a Brier-score-like loss. 

While survival analysis has received considerable attention in the machine learning community, the problem of competing risks is less well studied~\citep{wang2019machine} and even less applied~\citep{monterrubio2022review}, despite being central to medical applications. The existing methodologies to tackle competing risks rely on parametric assumptions, likelihood approximation, or optimise for a score that may misrepresent the true underlying survival distribution. This paper offers a novel competing risk model relying on constrained networks to obtain CIFs as a derivative instead of an integral. This approach results in the exact maximisation of the likelihood by leveraging automatic differentiation.

\section{Proposed approach}
This section formalises the problem of survival analysis and introduces the proposed model.

\subsection{Notation}
We model a population of the form $\{x_i, t_i, d_i\}_i$ with $x_i$ the covariates for patient $i$, $t_i\in \mathbb{R}^+$ the time of end of follow-up and $d_i\in [\![0, R]\!]$ its associated cause. If $d_i\in [\![1, R]\!]$, the patient left the study due to one of the $R$ considered risks. Otherwise, the patient is right-censored, i.e.,~the patient left the study for an \emph{unrelated} reason before any of the events of interest were observed. In this work, we focus on right-censoring, but the model can easily be extended to left-censoring. Note that we assume that experiencing one event precludes the observation of any other. 

\subsection{Survival quantities}
\paragraph{Single risk.} In settings with no competing risk,~i.e.,~$R = 1$, one aims to estimate the \emph{survival function} $S$, the probability of not observing the event of interest before time $t$, i.e.: $$S(t|x) := \mathbb{P}(T \geq t|x)$$

Equivalently, one aims to estimate the \emph{cumulative hazard function} $\Lambda(t|x)$ related to $S$ as follows: $$S(t|x) := \exp{\left[-\Lambda(t|x)\right]} = \exp{\left[-\int_0^t\lambda(u|x)du\right]}$$
where $\lambda(t|x) = \lim_{\delta t \to 0} \frac{\mathbb{P}(t < T < t +\delta t,| T \geq t, x)}{\delta t}$ is the instantaneous hazard of observing the event of interest, assuming no previous event(s).

Estimating this quantity may rely on maximising the likelihood of the observed data. The assumption of non-informative censoring,~i.e.,~event and censoring times are independent given the covariates, is necessary to express the likelihood. Specifically, each patient $i$ with an observed event contributes to the likelihood, the probability of experiencing the event at $t_i$ without previous events, i.e., $\lambda(t_i|x_i) S(t_i|x_i)$. The likelihood associated with each censored patient is the probability of not experiencing the event until $t_i$,~i.e.,~$S(t_i|x_i)$. This results in the following log-likelihood:

\begin{align}
    \label{eq:likelihood:single}
    l =  \sum_{i, d_i \neq 0} \log \lambda(t_i | x_i) - \sum_{i} \Lambda(t_i | x_i)
\end{align}

\begin{figure*}[hbt!]
    \centering
    \includegraphics[width=0.9\textwidth]{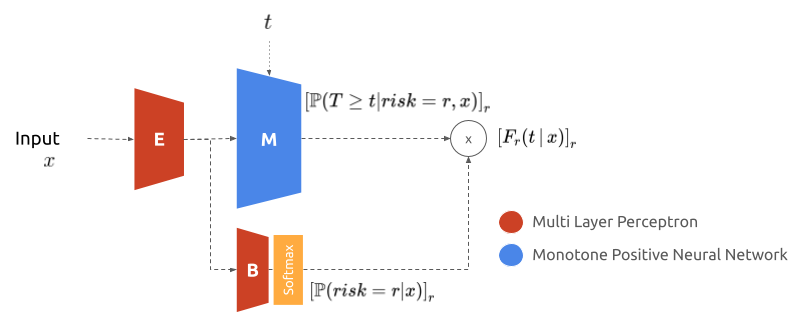}
    \caption{Neural Survival Analysis Architecture. \textit{$E$ embeds the covariate(s) $x$, which are then inputted in the monotonic networks $M$ and balancing network $B$ to estimate the CIFs.}}
    \label{fig:nfg}
\end{figure*}

\paragraph{Competing risks.} In the context of competing risks $R > 1$, a patient may leave a study for reasons correlated with the event of interest. Practitioners often consider these events as censoring and rely on single-risk models. However, this practice breaks the common assumption of non-informative censoring and results in misestimation of the survival function. When other events may be observed, $S(t \mid x)$ is defined as the probability of observing none of the competing risks before time $t$, i.e.: $$S(t|x) = 1 - \sum_{r \in [\![1, R]\!]} F_r(t|x)$$ where $F_r$, the \emph{Cumulative Incidence Function} (CIF) for the event $r$ denotes the probability of observing the event $r$ before time $t$ without prior occurrence of any competing event(s),~i.e.: 
\begin{align}
    F_r(t|x) = \mathbb{P}(T < t, \text{risk} = r| x)
\end{align}
with $T$, the random variable denoting the time of observation of any event. Note that the CIF can be expressed as an integral of observing the event in an infinitesimal interval given that no other event was observed until $t$: 
\begin{align}
    \label{eq:cif}
    F_r(t|x) = \int_0^t \lambda_r(u|x) e^{- \int_0^t \sum_r  \lambda_r(s) ds} du
\end{align}
with $\lambda_r(t|x) = \lim_{\delta t \to 0} \frac{\mathbb{P}(t < T < t +\delta t,\;\text{risk} = r | T \geq t,\;x)}{\delta t}$, the cause-specific hazard,~i.e., the instantaneous risk of observing the event $r$, with no other previous event.

A final quantity of interest is the cause-specific survival $S_r(t|x)$ that expresses the probability of not observing a given outcome $r$ by time $t$, i.e., 
\begin{align*}
    S_r(t|x) &= \mathbb{P}((T \geq t)\;\cup\;(T < t, \text{risk} \neq r)|x) \\
             &= 1 - F_r(t|x)
\end{align*}

Similar to the single-risk settings, we maximise the likelihood to estimate $F_r$. Importantly, we assume non-informative censoring \emph{once controlled} on all identified competing risks. While this assumption is more likely to hold once all competing risks are accounted for, practitioners suspecting its implausibility should perform sensitivity analysis for this assumption~\citep{jackson2014relaxing}. Under this assumption, the likelihood can be expressed analogously to~\eqref{eq:likelihood:single}: patients with an observed event contribute to the likelihood as the probability of observing the event $d_i$ at $t_i$ without observing any events until $t_i$, i.e., $\lambda_r(t_i|x_i) S(t_i|x_i)$. This quantity is the partial derivative of $F_r$ with respect to $t$ evaluated at $t_i$. Remaining censored patients influence the likelihood as the probability of observing no event until $t_i$,~i.e.,~$S(t_i|x_i)$. The competing risks log-likelihood can, therefore, be expressed as:

\begin{align}
    \label{eq:loglikelihood}
    l =  \sum_{r \in [\![1, R]\!]} &\sum_{i, d_i = r} \log \frac{\partial F_r(u|x_i)}{\partial u}\bigg|_{u = t_i} \\\nonumber+ &\sum_{i, d_i = 0} \log [1 - \sum_r F_r(t_i|x_i)]
\end{align}

One may extend existing models to the competing risks setting by performing the integration in~\eqref{eq:cif}. For instance, the cause-specific Cox model~\citep{prentice1978analysis} consists of Cox models independently trained on each risk,~i.e.,~treating all other outcomes as censored. Then one evaluates the CIF through~\eqref{eq:cif} using the estimated hazards. However, this staged modelling does not jointly consider the outcomes and may misestimate the covariate effects~\citep{van2018different}. Fine-Gray~\citep{fine1999proportional} overcomes this issue by directly modelling the sub-distribution hazards $h_r(t | x) = \lim_{\delta t \to 0} \frac{\mathbb{P}(t < T < t +\delta t,\;\text{risk} = r | (T \geq t)\;\cup\;(T < t\;\cap\;\text{risk} \neq r),\;x)}{\delta t}$, relying on a proportionality assumption of these quantities.

Likewise, one can extend machine learning architecture to enable the integration of the CIF and maximise the associated likelihood in~\eqref{eq:loglikelihood}. However, in the absence of a closed-form expression, this would necessitate numerical integration. This approximation may impact performance with added computational costs for training and predictions. Integration is computationally expensive, whereas derivation can be computed exactly in one backward pass by automatic differentiation -- available in most machine learning libraries. Therefore, our approach reduces the computational cost of the likelihood estimation by modelling $F_r$ and differentiating it to obtain $\lambda_r S$, resulting in the exact computation of all the previously described quantities of interest.

\subsection{Architecture}
\label{sec:model}
Neural Fine-Gray, illustrated in Figure~\ref{fig:nfg}, aims to model $[F_r]_{r \in [\![1, R]\!]}$ without relying on numerical integration to tackle the problem of competing risks. We decompose $F_r$ as:
\begin{align*}
    F_r(t|x) &= \mathbb{P}(\text{risk }=r| x) \cdot \mathbb{P}(T \leq t | \text{risk }=r, x) \\
             &= B(E(x))_r \cdot [1 - \exp(- t \times M_r(t, E(x)))]
\end{align*}

\paragraph{Embedding network ($E$).} A first multi-layer perceptron $E$ with inter-layer dropout extracts an embedding $\tilde{x}$ from the covariates $x$.

\paragraph{Sub-distribution networks ($[M_r]_{r \in [\![1, R]\!]}$).} The embedding $\tilde{x}$ is inputted in $R$ positive monotonic networks $[M_r]_{r \in [\![1, R]\!]}$ representing a lifetime distribution conditioned on one risk $r$, through the relation $1 - \exp(-t \times M_r(t, \tilde{x})) = \mathbb{P}(T \leq t | x, \text{risk }= r)$. A \emph{positive monotonic neural network} is a network constrained to have its outcome monotonic and positive given its input (see~\cite{daniels2010monotone} for theoretical analysis and~\cite{lang2005monotonic} for proof of universal approximator). Enforcing these constraints may rely on different transformations of the neural networks' weights~\citep{omi2019fully, rindt2022survival, chilinski2020neural}. In our work, we enforce all the neural networks' weights to be positive through a square function and use a final \textit{SoftPlus} layer to fulfil these constraints. Enforcing positive weights ensures that the outcome increases with the time dimension $t$. Additionally, enforcing a smooth function ensures a low computational cost and stable optimisation. Note that for model flexibility, we used $R$ monotonic networks. We explore in Appendix~\ref{supp-app:mono} how using one network with $R$ outcomes would impact performance.

\paragraph{Balancing network ($B$).} A multi-layer perceptron $B$ with a final \textit{SoftMax} layer leverages $\tilde{x}$ to balance the probability of observing each risk $B(\tilde{x}) := [\mathbb{P}(\text{risk }= r | x)]_r$. This weighting ensures that the survival function is correctly specified,~i.e.,~$\sum_{r \in [\![1, R]\!]} F_r(t|x) \leq 1$.\\

The proposed approach directly models $F_r$ by multiplying the outputs of the distribution and balancing networks. Automatic differentiation of the model's output results in the derivative $\frac{\partial F_r(u|x_i)}{\partial u}\bigg|_{u = t_i}$. The model can then be trained end-to-end by maximising the \emph{exact} log-likelihood proposed in Equation~\eqref{eq:loglikelihood}. By jointly modelling the competing risks, this proposed model is reminiscent of the Fine-Gray approach. The following equation exhibits the link between sub-distribution hazards and CIFs, i.e., between the standard and neural Fine-Gray models:
$$h_r(t | x) = \frac{1}{1 - F_r(t | x)} \cdot \frac{\partial F_r(u | x)}{\partial u}\bigg|_{u = t}$$

\begin{remark}
\label{rmk:correction}
\cite{shchur2019intensity} raise a limitation of monotonic neural networks that may attribute non-null density to negative times,~i.e.,~$F_r(t = 0|x) \neq 0$. In contrast to~\cite{omi2019fully, rindt2022survival}, we model $\mathbb{P}(T \leq t | \text{risk }=r, x)$ as $1 - \exp(- t \times M_r(t, \tilde{x}))$ instead of $M_r(t, \Tilde{x})$ to address this issue.
\end{remark}

\begin{remark}
The proposed methodology is a generalisation of the survival model Sumo-Net~\citep{rindt2022survival} that estimates $S$ in the single-risk setting. If $R = 1$, then $F_r = 1 - S$ and $B_r = \mathbbm{1}$. In this context, the proposed approach results in Sumo-Net. Moreover, the architecture resembles the one proposed in DeSurv~\citep{danks2022derivative} while avoiding numerical integration.
\end{remark}

\begin{table*}[]
    \begin{tabular}{c|ccccc}
    Dataset    & Observations & Features & Primary & Competing risk & Censored \\\midrule
    PBC        & 312                 & 25              & Death (44.87 \% ) & Transplant (9.29 \%)   & 45.83 \%       \\
    Framingham & 4,434               & 18              & CVD (26.09 \%) & Death (17.75 \%)         & 56.16 \%        \\
    Synthetic  & 30,000              & 12              & * (25.33 \%) &  * (24.67 \%)               & 50.00 \%    \\
    SEER       & 658,354             & 23              & BC (16.51 \%) & CVD (5.69 \%)    & 77.80 \%       \\
      
    \end{tabular}
    \caption{Datasets characteristics}
    \label{table:datasets:summary}
\end{table*}

\subsection{Computational complexity}
\label{sec:computationalcost}
Our modelling choices result in the exact computation of the likelihood. However, the other methodologies relying on integral approximation and outcome discretisation converge towards $F_r$ in the upper limit,~i.e., when increasing the number of point estimates, or using a finer discretisation. One may therefore question the advantage of the proposed methodology. In this section, we compare the complexity in estimating the CIF and likelihood for DeSurv~\citep{danks2022derivative}, the closest method to our proposed model, and NeuralFG. 

\paragraph{DeSurv}~\citep{danks2022derivative}. This approach models $F_r(t | x)$ as $\text{Tanh}(v(x, t))$ with $v$ being the solution to the ODE defined as $\frac{\partial v(x, u)}{\partial u}\bigg|_{u = t} = g(x, t)$ and $v(x, 0) = 0$ with $g$, a neural network. For efficiency, the authors propose a Gauss-Legendre quadrature to solve the ODE and obtain $v$. This approximation necessitates $n$ evaluations of $g$ at defined times $[t_j(t)]_{j \in [\![1, n]\!]}$ weighted by the associated $[w_j]_{j \in [\![1, n]\!]}$  (see~\cite{press2007numerical} for a detailed description of Gauss-Legendre quadrature). Each forward pass estimates $\frac{\partial v(x, u)}{\partial u}\bigg|_{u = t_j(t)}$ at the points used to approximate the integral, then $$\Hat{F}_r(t | x) = \text{Tanh}\left(\frac{t}{2}\sum_{j \in [\![1, n]\!]} w_j g(x, \frac{t}{2} t_j(t))\right)$$

\paragraph{DeSurv's computational cost.} Computation of $F_r$ relies on $n$ forward passes through the network. Moreover, the estimation of $\frac{\partial \Hat{F}_r(u|x_i)}{\partial u}\bigg|_{u = t_i}$ necessary to compute the competing risk likelihood is $g(x, t_i) (1 - \text{Tanh}(\Hat{F}_r(t_i | x))^2)$, ~i.e.,~$n + 1$ forward passes. The likelihood has a $\mathcal{O}(nN)$ computational complexity with $N$ the number of patients in the study. 

\paragraph{NeuralFG's computational cost.} $F_r$ is estimated in one forward pass and $\frac{\partial \Hat{F}_r(u|x_i)}{\partial u}\bigg|_{u = t_i}$ in one backward pass. Assuming the same computational cost for forward and backward passes, the likelihood estimation has a $\mathcal{O}(2N)$ complexity. Our proposed methodology, therefore, presents more than an $n/2$ computational gain compared to DeSurv in estimating the likelihood used for training, and an $n$ gain in inferring $F_r$. 

\section{Experiments}
This section introduces the datasets and experimental settings.

\subsection{Datasets}
We explore the model performance on four datasets with competing risks:
\begin{itemize}
    \item PBC~\citep{therneau2000cox} comprises 25 covariates in 312 patients over a 10-year randomised control trial to measure the impact of D-penicillamine on Primary Biliary Cholangitis (PBC). Death on the waiting list is the primary outcome with transplant being a competing risk. 
    \item Framingham~\citep{kannel1979diabetes} is a cohort study gathering 18 longitudinal measurements on male patients over 20 years. Our analysis focuses on the first observed covariates of 4,434 patients to model cardiovascular disease (CVD) risk. Death from other causes is treated as a competing risk. 
    \item Synthetic~\citep{lee2018deephit}, this dataset consists of 30,000 synthetic patients with 12 covariates following exponential event time distributions, non-linearly dependent on the covariates. 
    \item SEER\footnote{\url{https://seer.cancer.gov/}}: the Surveillance, Epidemiology, and End Results Program gathers covariates and outcomes of patients diagnosed with breast cancer between 1992 and 2017. Following the preprocessing proposed by~\cite{lee2018deephit, danks2022derivative}, we select 658,354 patients and 23 covariates describing the patient demographics and disease characteristics at diagnosis. Death from breast cancer (BC) is our primary outcome, with CVD, a competing risk. 
\end{itemize}

Table~\ref{table:datasets:summary} summarises the datasets' characteristics with the respective proportion of outcome and censoring.

\begin{table*}[!htb]
    \centering
        \centerline{\begin{tabular}{c|c|c|ccc||ccc}
            {} & \parbox[t]{2mm}{\multirow{2}{*}{\rotatebox[origin=c]{90}{Risk}}}  &\multirow{2}{*}{Model} & \multicolumn{3}{c||}{C-Index \textit{(Larger is better)}} & \multicolumn{3}{c}{Brier Score \textit{(Smaller is better)}} \\
             &  &  &           $q_{0.25}$ &           $q_{0.50}$ &           $q_{0.75}$ &  $q_{0.25}$ &           $q_{0.50}$ &           $q_{0.75}$ \\\midrule
            \parbox[t]{2mm}{\multirow{6}{*}{\rotatebox[origin=c]{90}{PBC}}} 
              & \parbox[t]{2mm}{\multirow{6}{*}{\rotatebox[origin=c]{90}{Death}}}
              & \textbf{NeuralFG} & 0.810 (0.079) & 0.795 (0.114) & 0.762 (0.123) & 0.099 (0.028) & 0.140 (0.020) & 0.169 (0.050) \\
              && DeepHit & 0.822 (0.099) & 0.844 (0.036) & 0.782 (0.033) & \textit{0.090} (0.030) & 0.132 (0.013) & 0.180 (0.021) \\
              && DeSurv & 0.821 (0.089) & 0.837 (0.050) & 0.815 (0.068) & \textbf{0.088} (0.022) & 0.113 (0.011) & \textbf{0.136} (0.047) \\
              && DSM & \textbf{0.867} (0.065) & \textbf{0.864} (0.037) & \textbf{0.828} (0.052) & 0.091 (0.039) & 0.124 (0.015) & 0.161 (0.022) \\
              && Fine-Gray & 0.831 (0.136) &    \textit{0.852} (0.045) &    \textit{0.816} (0.059) & 0.091 (0.042) &    \textit{0.103} (0.009) &    0.150 (0.038)\\
              && CS Cox & \textit{0.833} (0.125) &    0.851 (0.040) &    0.811 (0.065) & 0.091 (0.038) &    \textbf{0.102} (0.008) &    \textit{0.148} (0.038)\\\midrule
              
            \parbox[t]{2mm}{\multirow{6}{*}{\rotatebox[origin=c]{90}{Framingham}}} & \parbox[t]{2mm}{\multirow{6}{*}{\rotatebox[origin=c]{90}{CVD}}}
              & \textbf{NeuralFG} &  \textbf{0.872} (0.024) &  \textbf{0.812} (0.029) &  \textbf{0.782} (0.018) &  \textit{0.050} (0.003) &  \textbf{0.095} (0.010) &  \textbf{0.128} (0.004) \\
              && DeepHit & 0.855 (0.026) & 0.781 (0.026) & 0.743 (0.014) & 0.053 (0.003) & 0.102 (0.007) & 0.141 (0.002) \\
              && DeSurv &  \textbf{0.872} (0.027) & \textit{0.807} (0.031) & 0.775 (0.022) & \textbf{0.049} (0.005) & \textbf{0.095} (0.009) & \textit{0.129} (0.003) \\
              && DSM & \textit{0.866} (0.023) & 0.806 (0.023) & \textit{0.778} (0.014) & 0.057 (0.005) & 0.104 (0.006) & 0.141 (0.002) \\
              && Fine-Gray & 0.842 (0.025) &  0.794 (0.024) &  0.772 (0.015) &0.057 (0.006) &  0.099 (0.007) &  0.131 (0.003)\\
              && CS Cox & 0.845 (0.020) &  0.798 (0.022) &  0.774 (0.015) & 0.056 (0.006) &  \textit{0.098} (0.007) &  0.131 (0.003)\\\midrule
  
            \parbox[t]{2mm}{\multirow{6}{*}{\rotatebox[origin=c]{90}{Synthetic}}} & \parbox[t]{2mm}{\multirow{6}{*}{\rotatebox[origin=c]{90}{1}}}
              & \textbf{NeuralFG} & \textit{0.791} (0.013) & \textit{0.754} (0.013) & \textbf{0.715} (0.011) &\textbf{0.068} (0.003) & \textit{0.125} (0.004) & \textbf{0.192} (0.005) \\
              && DeepHit &  0.783 (0.012) & 0.747 (0.013) & \textit{0.714} (0.008) & 0.079 (0.003) & 0.136 (0.002) & \textit{0.212} (0.003) \\
              && DeSurv & \textbf{0.793} (0.013) & \textbf{0.756} (0.014) & \textit{0.714} (0.014) & \textbf{0.068} (0.002) &\textbf{0.124} (0.004) & \textbf{0.192} (0.004) \\
              && DSM & 0.776 (0.013) & 0.742 (0.013) & 0.710 (0.013) & \textit{0.073} (0.002) & 0.139 (0.002) & 0.220 (0.003) \\
              && Fine-Gray & 0.611 (0.014) &  0.587 (0.007) &  0.568 (0.009)& 0.078 (0.002) &  0.159 (0.003) &  0.241 (0.002)\\
              && CS Cox &0.609 (0.015) &  0.586 (0.006) &  0.568 (0.009)&0.078 (0.002) &  0.159 (0.003) &  0.240 (0.002)\\\midrule

            \parbox[t]{2mm}{\multirow{6}{*}{\rotatebox[origin=c]{90}{SEER}}} & \parbox[t]{2mm}{\multirow{6}{*}{\rotatebox[origin=c]{90}{BC}}}
              & \textbf{NeuralFG} & \textit{0.893} (0.002) & \textit{0.855} (0.001) & \textit{0.815} (0.001) &\textbf{0.038} (0.000) & \textbf{0.069} (0.001) & \textbf{0.101} (0.000) \\
              && DeepHit &  \textbf{0.899} (0.002) & \textbf{0.860} (0.001) & \textbf{0.818} (0.001) &\textbf{0.038} (0.000) & \textit{0.070} (0.000) & \textit{0.102} (0.001) \\
              && DeSurv &  0.892 (0.003) & 0.852 (0.002) & 0.813 (0.001) & \textbf{0.038} (0.000) & \textit{0.070} (0.000) & \textit{0.102} (0.001) \\
              && DSM & 0.884 (0.001) & 0.842 (0.002) & 0.805 (0.002) & \textit{0.039} (0.000) & 0.076 (0.001) & 0.112 (0.000)\\
              && Fine-Gray & 0.836 (0.003) &  0.786 (0.003) &  0.742 (0.002) &0.043 (0.001) &  0.081 (0.000) &  0.118 (0.000) \\
              && CS Cox &0.837 (0.003) &  0.786 (0.003) &  0.742 (0.002)& 0.042 (0.001) &  0.081 (0.000) &  0.118 (0.000) \\
        \end{tabular}}
    \caption{Comparison of model performance by means (standard deviations) across 5-fold cross-validation. Best performances are in \textbf{bold}, second best in \textit{italics}. \textit{NeuralFG is the model introduced in this paper.}}
    \label{tab:result}
\end{table*}

\subsection{Baseline models}
The proposed Neural Fine-Gray~(\textbf{NeuralFG}) was compared against six strategies. First, we considered the well-established cause-specific Cox model (\textbf{CS Cox}~\cite{prentice1978analysis}) and \textbf{Fine-Gray} model~\citep{fine1999proportional} with a linear parametric form for the covariate effect. The cause-specific Cox model models each cause independently using a Cox proportional-hazards model, while Fine-Gray models the sub-hazard functions assuming proportional sub-hazards. 

Thereafter, we compare state-of-the-art competing risk survival neural networks proposed in the machine learning literature. First, Deep Survival Machine~(\textbf{DSM},~\cite{nagpal2020deep}) consists of a mixture of Weibull distributions parameterised by neural networks. Each point is then assigned to these distributions through an assignment network. Using parametric distributions results in a closed-form likelihood in the competing risks setting. \textbf{DeepHit}~\citep{lee2018deephit} discretises the survival horizon and leverages a multi-head network to associate each patient to the interval corresponding to its observed event time and type. Each head of the network is associated with one cause as in the proposed NeuralFG. The time-discretisation results in a discrete likelihood further penalised by a C-index-like regularisation for model training. Closer to our work, \textbf{DeSurv}~\citep{danks2022derivative} approaches $F_r$ as the solution to an ODE.

\begin{table*}[!htb]
    \centering
        \centerline{\begin{tabular}{c|c|ccc||ccc}
            \multirow{2}{*}{Death}  &\multirow{2}{*}{Model} & \multicolumn{3}{c||}{C-Index \textit{(Larger is better)}} & \multicolumn{3}{c}{Brier Score \textit{(Smaller is better)}} \\
              &  &           $q_{0.25}$ &           $q_{0.50}$ &           $q_{0.75}$ &  $q_{0.25}$ &           $q_{0.50}$ &           $q_{0.75}$ \\\midrule
                \multirow{2}{*}{CVD}
                  & \textbf{Competing} & \textbf{0.872} (0.024) &  \textbf{0.812} (0.029) &  \textbf{0.782} (0.018) &  \textbf{0.050} (0.003) &  \textbf{0.095} (0.010) &  \textbf{0.128} (0.004) \\
                  & {Non-Competing} & 0.862 (0.029) &  0.807 (0.032) &  0.780 (0.020) &  0.053 (0.004) &  0.099 (0.011) &  0.129 (0.005) \\ \midrule
                \multirow{2}{*}{Death}
                  & \textbf{Competing} &  \textbf{0.745} (0.055) &  0.717 (0.038) &  0.713 (0.022) &  \textbf{0.027} (0.003) &  \textbf{0.070} (0.004) &  0.112 (0.005) \\
                  & {Non-Competing} &  0.741 (0.053) &  \textbf{0.718} (0.045) &  \textbf{0.719} (0.025) &  \textbf{0.027} (0.003) &  0.071 (0.002) &  \textbf{0.109} (0.004) \\
        \end{tabular}}
    \caption{Modelling competing risk - means (standard deviations) across the 5-fold cross-validation.}
    \label{tab:comparison}
\end{table*}

\subsection{Experimental settings}
\label{sec:exp}
The analysis relies on 5-fold cross-validation with 10\% of each training set left aside for hyper-parameter tuning. Random search is used on the following grid over 100 iterations: learning rate ($10^{-3}$ or $10^{-4}$), batch size ($100$, $250$, except for SEER: $1,000$ or $5,000$), dropout rate ($0$, $0.25$, $0.5$ or $0.75$), number of layers ($[\![1, 4]\!]$) and nodes ($25$ or $50$). All activation functions are fixed to \textit{Tanh} to ensure a properly defined derivative -- note that any $\mathcal{C}^1$ activation would work. All models are optimised using an Adam optimiser~\citep{kingma2014adam} over $1,000$ epochs, with an early stopping criterion computed on a 10\% left-aside subset of the training set. 

Other methods are optimised over the same grid (if applicable). Additionally, we explore both Log-Normal and Weibull distributions for DSM and use $10,000$ warm-up iterations to estimate the parametric form closest to the average survival as proposed in the original paper~\citep{nagpal2020deep}. For DeSurv, we followed the original paper's recommendation of a 15-degree Gauss-Legendre quadrature to estimate the CIFs. In Appendix~\ref{supp-app:comparison}, we further investigate how increasing the number of point estimates impacts performance. We use a similar approximation for DeepHit with a 15-split time discretisation. Finally, for a fair comparison, we double the number of possible layers for architectures without embedding networks.

\subsection{Evaluation metrics}
As per current practice in survival literature, we used the time-dependent Brier score~\citep{graf1999assessment} to quantify calibration, and the C-index~\citep{antolini2005time} for discrimination at the dataset-specific 0.25, 0.5 and 0.75 quantiles of the uncensored population event times (See Appendix~\ref{supp-sec:characteristics} for data characteristics, \ref{supp-app:metric} for further description of the metrics and \ref{supp-app:cumulative} for the cumulative version of these metrics). Means and standard deviations are computed over the 5 folds of cross-validation.

\section{Results}
Table~\ref{tab:result} summarises the calibration and discriminative performance of the analysed models on the primary outcome (see Appendix~\ref{supp-app:moremetric} for the performances on the competing risk). 

\subsection{Model's strengths}
NeuralFG demonstrates lower or equal Brier scores than other state-of-the-art machine learning models across the majority of datasets and time horizons.

While DSM presents good discriminative performances, this edge is not reflected in its calibration. This observation indicates that parametric assumptions may result in estimated survival functions discriminative of the outcome but further from the underlying survival distribution. Deep-Hit penalisation results in better C-Index values but hurts model calibration, with misaligned discrimination and calibration throughout the different datasets. 

Finally, performances are comparable to DeSurv. However, DeSurv's likelihood approximation multiplies its computational cost by the numerical integration complexity (see Appendix~\ref{supp-app:speed} for a comparison of training speed on the Framingham dataset). NeuralFG, therefore, achieves state-of-the-art performance while avoiding computationally-expensive approximations.

\subsection{Model's limitations}
The proposed methodology has lower performance on the PBC dataset, which notably comprises a limited amount of data. In small-data settings, practitioners should prefer simpler models to avoid overfitting.

For instance, the linear Fine-Gray and CS Cox models result in competitive performances on PBC. However, this linearity assumption hurts performance under more complex covariate effects as in the SEER and Synthetic datasets. Note that leveraging domain expertise could enhance performance through the addition of interactions and the use of alternative models. However, these approaches deviate from the automated discovery of interactions facilitated by neural networks. Similarly, the parametric assumption of DSM results in the best discrimination in PBC, but it under-performs under more complex survival distributions. 

Furthermore, the DeSurv model performs better than the proposed methodology on PBC. This may reflect that approximating the likelihood can regularise model training, which is beneficial in the context of small data.


\subsection{Modelling vs ignoring competing risks}
This last section explores the importance of modelling competing risks in the Framingham dataset. First, we present the performance differences between the proposed model in comparison to the same architecture maximising the cause-specific likelihoods. Then, we explore which subgroups of the population most benefit from this modelling. Finally, we study how guidelines would differ under the proposed NeuralFG and its non-competing alternative.

\paragraph{Why account for competing risks?} To measure how modelling competing risks impacts performance, while ensuring the \emph{same number of parameters}, we propose to use the same architecture presented in Section~\ref{sec:model} whilst maximising the sum of the cause-specific likelihoods,~i.e.:
$$l = \sum_r \left[\sum_{i, d_i = r} \log \lambda_r(t_i | \Tilde{x}_i) - \sum_{i} \Lambda_r(t_i | \Tilde{x}_i)\right]$$
Each monotonic network, therefore, models the cumulative hazard function for risk $r$, $\Lambda_r$, by maximising the likelihood of one cause whilst considering the rest of the population as censored, relying on a shared embedding $\Tilde{x}$. Automatic differentiation outputs $[\lambda_r]_{r \in [\![1, R]\!]}$. Table~\ref{tab:comparison} summarises the discrimination and calibration differences in the non-competing survival $e^{-\Lambda_r(t | x)}$ obtained with this model and the previously described NFG's cause-specific survival $1 - F_r(t | x)$. Note how modelling competing risks significantly improves performance for the primary outcome of interest, CVD, without significant differences for the competing risk. Since patients who die from other causes during the study period do not present the same risk of CVD as patients remaining in the study, not accounting for all-cause mortality results in a misestimation of CVD risk.

\paragraph{Who may benefit?} One can explore which subgroups benefit the most from modelling competing risks. Intuitively, patients who are the most likely to suffer from competing risks may benefit the most from this modelling. Table~\ref{tab:age} illustrates this with older patients benefiting the most from modelling death as a competing risk.

\begingroup
\setlength{\tabcolsep}{1pt}
\begin{table}[!ht]
    \centering
        \centerline{\begin{tabular}{c|ccc}
            \multirow{2}{*}{Age}  & \multicolumn{3}{c}{Brier Score Difference}\\
               &     $q_{0.25}$ &           $q_{0.50}$ &           $q_{0.75}$ \\\midrule
                $<40$ &  -0.000 (0.000) &  -0.001 (0.002) &   0.000 (0.005) \\
                40-50 &  -0.001 (0.001) &  -0.002 (0.003) &  -0.002 (0.001) \\
                50-60 &  \textit{-0.003} (0.005) &  \textit{-0.004} (0.003) &  \textit{-0.006} (0.007) \\
                60+   &  \textbf{-0.013} (0.011) &  \textbf{-0.022 (0.018)} &  \textbf{-0.007} (0.024) \\
        \end{tabular}}
    \caption{Calibration differences - Means and standard deviations over 5-fold cross-validation. \textit{Larger negative values correspond to better calibration for the competing risk model.}}
    \label{tab:age}
\end{table}
\endgroup

\begingroup
\setlength{\tabcolsep}{3pt}
\begin{table*}[!ht]
    \centering
    \caption{Reclassification matrices between competing and non-competing risk scores for patients older than 50. \textit{Red (resp. blue) shows when the competing risks score is less aligned with the 10-year observed outcome than the non-competing model (resp. more aligned). Note that censored patients are ignored.}}
    \label{tab:risk}
    \subtable[Patients \textit{with no event} in the 10-year follow-up.]{
      \centering
      \vspace{100pt}
        \begin{tabular}{cc|ccc|c}
           &                 & \multicolumn{3}{c|}{Non - Competing}&\\
             & Risk&   Low &  Inter. &  High & Total  \\\midrule
            \parbox[t]{2mm}{\multirow{3}{*}{\rotatebox[origin=c]{90}{Comp.}}} &
             Low           &  502 &     \green{228} &    \green{23} &        753 \\
            & Intermediate        &    \red{2} &     189 &   \green{229} &        420 \\
            & High          &    \red{0} &       \red{9} &   314 &        323 \\\midrule
            &  Total             &  504 &     426 &   566 &       1496 \\
        \end{tabular}}\qquad
    \subtable[Patients \textit{with} an observed event during the 10-year follow-up.]{
      \centering
        \begin{tabular}{cc|ccc|c}
           &                 & \multicolumn{3}{c|}{Non - Competing}&\\
             & Risk&   Low &  Inter. &  High & Total  \\\midrule
            \parbox[t]{2mm}{\multirow{3}{*}{\rotatebox[origin=c]{90}{Comp.}}} &
             Low           &   23 &      \red{28} &     \red{5} &        56 \\
            & Intermediate        &    \green{1} &      37 &    \red{41} &         79 \\
            & High          &    \green{2} &       \green{4} &   248 &        254 \\\midrule
            &    Total           &  26 &      69 &   294 &        389 \\
        \end{tabular}} 
\end{table*}
\endgroup

\paragraph{What is the impact on medical practice?} The Framingham dataset was used to model the eponymous 10-year cardiovascular disease (CVD) risk score~\citep{wilson1998prediction}. This score guides clinical practice in preventatively treating patients, usually with a combination of cholesterol-lowering therapy,~e.g.,~statins, and holistic treatment of other CVD risk factors~\citep{bosomworth2011practical}. To minimise overtreatment and adverse side effects, accurate risk estimates are critical for targeting the population most at risk so as to maximise the benefit-risk ratio~\citep{mangione2022statin}. However, the original Framingham score relies on a non-competing risk model~\citep{mangione2022statin, van2014performance}. 

Clinical treatment often relies on a discretisation of this risk~\citep{bosomworth2011practical}: low, intermediate and high risk, at $< 10\%$, $10 - 20\%$ and $> 20\%$ chance, respectively, of observing a CVD event in the following 10 years. Current guidelines in the United States suggest placing all patients with $\geq 10\%$ risk on cholesterol-lowering drugs~\citep{mangione2022statin}. Furthermore, in the US alone, several million patients are on these medications~\citep{wall2018vital}. Therefore, even modest shifts in patient risk classification could, at scale, amount to considerable numbers either inappropriately receiving preventative treatment or inappropriately receiving none. To demonstrate how considering competing risks can fundamentally alter such risk profiling, we present in Table~\ref{tab:risk} the reclassification matrices of risk levels given competing and non-competing NeuralFG differentiated by observed outcomes for patients aged 50 or over. For instance, note that 251 deemed intermediate-to-high risk by the non-competing risks model are reclassified as lower risk by the competing-risks model, who, in turn, could have avoided the initiation of therapy. These results echo the medical literature's findings of risk misestimation due to the non-consideration of competing risks in this risk score~\citep{lloyd2004framingham, van2014performance}. More accurate simulations to estimate the potential lives saved and harmed through such reclassification is beyond the scope of this article but could provide insight into the possible consequences of considering competing risks. In summary, using a non-competing risk score would have important clinical consequences of over- and under-treatment~\citep{schuster2020ignoring}. More predictive models accounting for competing risks must be preferred to ensure better care.

\section{Conclusion}
This work provides a solution to address competing risks that preclude the observation of the outcome of interest, often present in medical applications. We introduce Neural Fine-Gray, a monotonic neural network architecture, to tackle the problem of competing risks in survival modelling. The model outputs the cumulative incidence functions and, consequently, allows the exact likelihood computation. Importantly, this architecture choice achieves competitive performance while avoiding the parametric assumptions or computationally expensive approximations made by state-of-the-art survival neural networks. Further analysis of the Framingham dataset contributes to the literature, inviting practitioners to use competing-risk modelling in risk score development for improved care~\citep{abdel2018importance, austin2016developing, lloyd2004framingham, schuster2020ignoring}.

Our future work will (i)~extend this architecture to model other modalities such as time series as in~\cite{pmlr-v146-nagpal21a} and, (ii)~explore medically interpretable survival clusters as presented in~\cite{jeanselme2022neural, nagpal2022counterfactual}.

\acks{This work was supported by The Alan Turing Institute’s Enrichment Scheme and partially funded by UKRI Medical Research Council (MC\_UU\_00002/5 and MC\_UU\_00002/2).}
\clearpage
\bibliography{references}

\clearpage
\appendix
\section{Experiments}

\subsection{Datasets characteristics}
\label{supp-sec:characteristics}
Table~\ref{tab:quantiles} presents the times and observed outcomes corresponding to the different quantiles of the uncensored population used for evaluation, differentiated by datasets.

\begin{table}[!h]
    \centerline{\begin{tabular}{ccccc}
         &  &  \multicolumn{3}{c}{Quantiles}  \\
         &  &  $q_{0.25}$ &           $q_{0.50}$ &           $q_{0.75}$  \\\midrule
         \parbox[t]{2mm}{\multirow{4}{*}{\rotatebox[origin=c]{90}{PBC}}} & Time (years) & 3.19 & 4.95 & 7.45 \\\cdashline{2-5}
         & Censoring & 0.00 \% & 0.00\% & 11.54\% \\
         & Death & 12.82\% & 23.72\% & 32.69\%  \\
         & Transplant & 0.64\% & 3.21\% & 7.69\%\\\midrule

         \parbox[t]{2mm}{\multirow{4}{*}{\rotatebox[origin=c]{90}{Fram.}}} & Time (years) & 5.90 & 12.57 & 18.14 \\\cdashline{2-5}
         & Censoring & 0.00 \% & 0.00\% & 0.00\% \\
         & Death & 2.66\% & 7.40\% & 12.56\%  \\
         & CVD & 8.30\% & 14.52\% & 20.32\%\\\midrule

         \parbox[t]{2mm}{\multirow{4}{*}{\rotatebox[origin=c]{90}{Synth.}}} & Time & 3.00 & 11.00 & 30.00 \\\cdashline{2-5}
         & Censoring & 20.54\% & 35.32\% &  44.65\% \\
         & Cause 1 & 5.28\% & 12.43\% &  18.96\% \\
         & Cause 2 & 5.05\% & 12.21\% &  18.43\% \\\midrule

         \parbox[t]{2mm}{\multirow{4}{*}{\rotatebox[origin=c]{90}{SEER}}} & Time (years) & 1.67 & 4.00 & 8.08 \\\cdashline{2-5}
         & Censoring & 10.34\% & 22.20 \% &  39.59\% \\
         & BC & 4.53\% & 9.32\% &  13.37\% \\
         & CVD & 0.80\% & 1.76\% &  3.23\% \\
    \end{tabular}}
\caption{Observed outcomes of interest at the different evaluation horizons.}
\label{tab:quantiles}
\end{table}

\subsection{Evaluation metrics}
\label{supp-app:metric}

\paragraph{Time Dependent C-Index}
Time-dependent C-Index~\citep{antolini2005time} quantifies the model discrimination by comparing the ordering of the predicted survival probability for risk $r$ and the observed survival times, i.e.~it is an estimate of: 
\begin{align*}
    &\mathbbm{P}(\Hat{F}_r(t|x_i) > \Hat{F}_r(t|x_j)| d_i = r, t_i < t_j , t_i \leq t)\\
\end{align*}
This probability is approximated and weighted by the inverse probability $\omega(t_i)$ of censoring derived from a Kaplan-Meier estimator. 

\paragraph{Time Dependent Brier Score}
Time dependent Brier score~\citep{graf1999assessment} measures the model calibration for risk $r$, similarly corrected for censoring:
\begin{align*}
\text{BS}^r(t) = \frac{1}{n} \sum_i \big[& \omega(t_i) \mathbbm{1}_{i, d_i = r \wedge t_i \leq t} (1 - \hat{F}_r(t|x_i))^2 \\
+& \omega(t) \mathbbm{1}_{t_i > t} \hat{F}_r(t|x_i)^2\big]
\end{align*}
with $\mathbbm{1}$, the indicator function, $\hat{S}(t|x)$, the predicted survival probability at time $t$.

\subsection{Time specific results}
\label{supp-app:moremetric}

Tables~\ref{tab:res:cindex},~\ref{tab:res:auc} and~\ref{tab:res:brier} present the performance evaluated at the dataset-specific 0.25, 0.5 and 0.75 quantiles of the uncensored population event times through respectively C-index, ROC-AUC, and Brier score. The table echoes the same conclusions presented in the paper with competing or better than state-of-the-art performance. 

\begin{table*}[p]
    \centering
        \centerline{\begin{tabular}{c|c|ccc||ccc}
            {}   &\multirow{2}{*}{Model} & \multicolumn{3}{c||}{Primary risk} & \multicolumn{3}{c}{Competing risk} \\
              &  &           $q_{0.25}$ &           $q_{0.50}$ &           $q_{0.75}$ &  $q_{0.25}$ &           $q_{0.50}$ &           $q_{0.75}$ \\\midrule
            \parbox[t]{2mm}{\multirow{6}{*}{\rotatebox[origin=c]{90}{PBC}}} 
              & \textbf{NeuralFG} & 0.810 (0.079) & 0.795 (0.114) & 0.762 (0.123)& 0.799 (0.082) & 0.709 (0.309) & \textit{0.788} (0.145) \\
              & DeepHit & 0.822 (0.099) & 0.844 (0.036) & 0.782 (0.033)& 0.790 (0.044) & 0.614 (0.174) & 0.612 (0.095) \\
              & DeSurv&  0.821 (0.089) & 0.837 (0.050) & 0.815 (0.068)& 0.802 (0.123) & \textbf{0.781} (0.268) & \textbf{0.796} (0.153) \\
              & DSM & \textbf{0.867} (0.065) & \textbf{0.864} (0.037) & \textbf{0.828} (0.052) & 0.694 (0.224) & 0.721 (0.251) & 0.703 (0.175) \\
              & Fine-Gray & 0.831 (0.136) &    \textit{0.852} (0.045) &    \textit{0.816} (0.059) &\textbf{0.865} (0.087) &    0.686 (0.330) &    0.741 (0.123)\\
              & CS Cox & \textit{0.833} (0.125) &    0.851 (0.040) &    0.811 (0.065) & \textit{0.837} (0.022) &    \textit{0.734} (0.276) &    0.783 (0.118)\\\midrule
              
            \parbox[t]{2mm}{\multirow{6}{*}{\rotatebox[origin=c]{90}{Framingham}}} 
              & \textbf{NeuralFG} &   \textbf{0.872} (0.024) &  \textbf{0.812} (0.029) &  \textbf{0.782} (0.018) &\textbf{0.745} (0.055) & \textbf{0.717} (0.038) & \textit{0.713} (0.022) \\
              & DeepHit & 0.855 (0.026) & 0.781 (0.026) & 0.743 (0.014)& 0.713 (0.035) & 0.690 (0.030) & 0.693 (0.015) \\
              & DeSurv& \textbf{0.872} (0.027) & \textit{0.807} (0.031) & 0.775 (0.022)  &0.721 (0.036) & 0.706 (0.038) & 0.708 (0.028) \\
              & DSM & \textit{0.866} (0.023) & 0.806 (0.023) & \textit{0.778} (0.014)& 0.717 (0.064) & 0.709 (0.034) & 0.712 (0.021)\\
              & Fine-Gray & 0.842 (0.025) &  0.794 (0.024) &  0.772 (0.015) &0.729 (0.036) &  0.709 (0.040) &  0.710 (0.023) \\
              & CS Cox & 0.845 (0.020) &  0.798 (0.022) &  0.774 (0.015) & \textit{0.741} (0.050) &  \textit{0.712} (0.044) &  \textbf{0.715} (0.023)\\\midrule

            \parbox[t]{2mm}{\multirow{6}{*}{\rotatebox[origin=c]{90}{Synthetic}}} 
              & \textbf{NeuralFG} &  \textit{0.791} (0.013) & \textit{0.754} (0.013) & \textbf{0.715} (0.011)& \textit{0.801} (0.016) & \textit{0.755} (0.018) & \textit{0.714} (0.016) \\
              & DeepHit & 0.783 (0.012) & 0.747 (0.013) & \textit{0.714} (0.008) &0.792 (0.015) & 0.744 (0.015) & \textbf{0.715} (0.012) \\
              & DeSurv& \textbf{0.793} (0.013) & \textbf{0.756} (0.014) & \textit{0.714} (0.014) & \textbf{0.803} (0.015) & \textbf{0.756} (0.016) & 0.713 (0.015)\\
              & DSM &  0.776 (0.013) & 0.742 (0.013) & 0.710 (0.013)& 0.785 (0.019) & 0.742 (0.019) & 0.708 (0.020)\\
              & Fine-Gray & 0.611 (0.014) &  0.587 (0.007) &  0.568 (0.009) & 0.633 (0.014) &  0.593 (0.015) &  0.574 (0.015) \\
              & CS Cox &0.609 (0.015) &  0.586 (0.006) &  0.568 (0.009) & 0.630 (0.013) &  0.592 (0.014) &  0.573 (0.015) \\\midrule

            \parbox[t]{2mm}{\multirow{6}{*}{\rotatebox[origin=c]{90}{SEER}}} 
              & \textbf{NeuralFG} & \textit{0.893} (0.002) & \textit{0.855} (0.001) & \textit{0.815} (0.001) & 0.799 (0.010) & 0.782 (0.005) & \textit{0.758} (0.003)\\
              & DeepHit & \textbf{0.899} (0.002) & \textbf{0.860} (0.001) & \textbf{0.818} (0.001) &\textbf{0.824} (0.008) & \textbf{0.801} (0.005) & \textbf{0.770} (0.004)\\
              & DeSurv& 0.892 (0.003) & 0.852 (0.002) & 0.813 (0.001) & 0.811 (0.006) & \textit{0.788} (0.006) & 0.757 (0.004)\\
              & DSM & 0.884 (0.001) & 0.842 (0.002) & 0.805 (0.002) & \textit{0.813} (0.008) & 0.787 (0.004) & 0.755 (0.004)\\
              & Fine-Gray & 0.836 (0.003) &  0.786 (0.003) &  0.742 (0.002)& 0.757 (0.008) &  0.745 (0.005) &  0.727 (0.005)\\
              & CS Cox & 0.837 (0.003) &  0.786 (0.003) &  0.742 (0.002)&0.781 (0.010) &  0.759 (0.007) &  0.734 (0.006)\\
        \end{tabular}}
    \caption{Comparison of the \textbf{C-index} across 5-fold cross-validation. Best performances are in \textbf{bold}, second best in \textit{italics}. \textit{NeuralFG is the model introduced in this paper.}}
    \label{tab:res:cindex}
\end{table*}

\begin{table*}[p]
    \centering
        \centerline{\begin{tabular}{c|c|ccc||ccc}
            {}   &\multirow{2}{*}{Model} & \multicolumn{3}{c||}{Primary risk} & \multicolumn{3}{c}{Competing risk} \\
              &  &           $q_{0.25}$ &           $q_{0.50}$ &           $q_{0.75}$ &  $q_{0.25}$ &           $q_{0.50}$ &           $q_{0.75}$ \\\midrule
            \parbox[t]{2mm}{\multirow{6}{*}{\rotatebox[origin=c]{90}{PBC}}} 
              & \textbf{NeuralFG} & 0.822 (0.088) &    0.825 (0.145) &    0.809 (0.161)  &0.804 (0.097) &    0.741 (0.316) &    \textbf{0.842} (0.169)\\
              & DeepHit & 0.831 (0.104) &    0.876 (0.054) &    0.803 (0.057) &0.786 (0.045) &    0.610 (0.175) &    0.623 (0.102)\\
              & DeSurv& 0.823 (0.088) &    0.866 (0.065) &    \textbf{0.855} (0.100) & 0.807 (0.113) &    \textbf{0.795} (0.269) &    \textit{0.831} (0.174) \\
              & DSM & \textbf{0.876} (0.067) &    \textbf{0.900} (0.043) &    \textit{0.854} (0.062) & 0.707 (0.228) &    0.728 (0.242) &    0.695 (0.175) \\
              & Fine-Gray & 0.835 (0.136) &    \textit{0.887} (0.059) &    0.844 (0.089)&\textbf{0.871} (0.075) &    0.706 (0.336) &    0.754 (0.133) \\
              & CS Cox &\textit{0.839} (0.127) &    0.886 (0.056) &    0.843 (0.097)&\textit{0.843} (0.009) &    \textit{0.750} (0.272) &    0.798 (0.123)\\\midrule
              
            \parbox[t]{2mm}{\multirow{6}{*}{\rotatebox[origin=c]{90}{Framingham}}} 
              & \textbf{NeuralFG} &  \textbf{0.877} (0.025) &  \textbf{0.827} (0.028) &  \textbf{0.810} (0.016) & \textbf{0.752} (0.056) &  \textbf{0.736} (0.042) &  \textit{0.742} (0.024)\\
              & DeepHit & 0.860 (0.026) &  0.796 (0.024) &  0.770 (0.015)& 0.720 (0.034) &  0.708 (0.037) &  0.723 (0.018) \\
              & DeSurv& \textit{0.876} (0.028) &  \textit{0.821} (0.030) &  {0.803} (0.020) &0.728 (0.035) &  0.724 (0.043) &  0.736 (0.032) \\
              & DSM & 0.870 (0.023) &  0.819 (0.022) &  {0.803} (0.015) &0.722 (0.065) &  0.727 (0.039) &  0.741 (0.024) \\
              & Fine-Gray & 0.849 (0.027) &  0.812 (0.023) &  0.802 (0.015) & 0.736 (0.036) &  0.727 (0.044) &  0.739 (0.027)\\
              & CS Cox & 0.852 (0.022) &  0.816 (0.021) &  \textit{0.804} (0.015) & \textit{0.748} (0.051) &  \textit{0.730} (0.047) &  \textbf{0.745} (0.025)\\\midrule

            \parbox[t]{2mm}{\multirow{6}{*}{\rotatebox[origin=c]{90}{Synthetic}}} 
              & \textbf{NeuralFG} & \textit{0.814} (0.015) &  \textit{0.806} (0.012) &  \textbf{0.790} (0.015)& \textit{0.821} (0.021) &  \textit{0.804} (0.017) &  \textit{0.785} (0.015)  \\
              & DeepHit & 0.806 (0.016) &  0.803 (0.013) &  \textit{0.788} (0.015)&  0.814 (0.020) &  0.796 (0.015) &  \textbf{0.790} (0.013)\\
              & DeSurv& \textbf{0.817} (0.016) &  \textbf{0.809} (0.013) &  0.787 (0.017) & \textbf{0.824} (0.020) &  \textbf{0.805} (0.016) &  0.780 (0.013)\\
              & DSM & 0.800 (0.016) &  0.794 (0.013) &  0.782 (0.013)& 0.807 (0.023) &  0.790 (0.020) &  0.776 (0.022) \\
              & Fine-Gray &  0.603 (0.018) &  0.583 (0.008) &  0.562 (0.005) & 0.624 (0.014) &  0.585 (0.018) &  0.565 (0.018)\\
              & CS Cox & 0.601 (0.018) &  0.583 (0.008) &  0.562 (0.005)& 0.621 (0.013) &  0.584 (0.018) &  0.565 (0.018)\\\midrule

            \parbox[t]{2mm}{\multirow{6}{*}{\rotatebox[origin=c]{90}{SEER}}} 
              & \textbf{NeuralFG} & \textit{0.901} (0.001) &  \textit{0.868} (0.002) &  \textit{0.826} (0.001) & 0.804 (0.010) &  0.789 (0.006) &  0.761 (0.003) \\
              & DeepHit &  \textbf{0.907} (0.002) &  \textbf{0.874} (0.001) &  \textbf{0.835} (0.002)&\textbf{0.832} (0.008) &  \textbf{0.814} (0.006) &  \textbf{0.783} (0.004) \\
              & DeSurv& 0.899 (0.003) &  0.866 (0.002) &  0.825 (0.001)& 0.818 (0.006) &  \textit{0.798} (0.007) &  \textit{0.764} (0.004)\\
              & DSM & 0.891 (0.001) &  0.855 (0.002) &  0.815 (0.002) &  \textit{0.821} (0.009) &  0.796 (0.004) &  0.758 (0.005)\\
              & Fine-Gray &0.840 (0.003) &  0.799 (0.003) &  0.757 (0.002) & 0.760 (0.009) &  0.749 (0.005) &  0.736 (0.005) \\
              & CS Cox &0.841 (0.003) &  0.799 (0.003) &  0.758 (0.002)& 0.785 (0.010) &  0.766 (0.007) &  0.745 (0.006)\\
        \end{tabular}}
    \caption{Comparison of the \textbf{time-dependent AUC} across 5-fold cross-validation. Best performances are in \textbf{bold}, second best in \textit{italics}. \textit{NeuralFG is the model introduced in this paper.}}
    \label{tab:res:auc}
\end{table*}

\begin{table*}[p]
    \centering
        \centerline{\begin{tabular}{c|c|ccc||ccc}
            {}   &\multirow{2}{*}{Model} & \multicolumn{3}{c||}{Primary risk} & \multicolumn{3}{c}{Competing risk} \\
              &  &           $q_{0.25}$ &           $q_{0.50}$ &           $q_{0.75}$ &  $q_{0.25}$ &           $q_{0.50}$ &           $q_{0.75}$ \\\midrule
            \parbox[t]{2mm}{\multirow{6}{*}{\rotatebox[origin=c]{90}{PBC}}} 
              & \textbf{NeuralFG} & 0.099 (0.028) &    0.140 (0.020) &    0.169 (0.050) & \textit{0.018} (0.001) &    0.036 (0.015) &    0.092 (0.017) \\
              & DeepHit &  \textit{0.090} (0.030) &    0.132 (0.013) &    0.180 (0.021) & \textit{0.018} (0.002) &    0.039 (0.020) &    0.100 (0.010)\\
              & DeSurv& \textbf{0.088 }(0.022) &    0.113 (0.011) &    \textbf{0.136} (0.047)& 0.019 (0.001) &    \textbf{0.031} (0.011) &    \textbf{0.087} (0.020)\\
              & DSM & {0.091} (0.039) &    0.124 (0.015) &    0.161 (0.022) & \textbf{0.017} (0.000) &    \textit{0.035} (0.018) &    0.099 (0.017)\\
              & Fine-Gray &{0.091} (0.042) &    \textit{0.103} (0.009) &    0.150 (0.038)& \textbf{0.017} (0.000) &    0.041 (0.017) &    \textit{0.091} (0.017) \\
              & CS Cox &{0.091} (0.038) &    \textbf{0.102} (0.008) &    \textit{0.148} (0.038) & \textit{0.018} (0.000) &    0.038 (0.018) &    \textbf{0.087} (0.018)\\\midrule
              
            \parbox[t]{2mm}{\multirow{6}{*}{\rotatebox[origin=c]{90}{Framingham}}} 
              & \textbf{NeuralFG} &\textit{0.050} (0.003) &  \textbf{0.095} (0.010) &  \textbf{0.128} (0.004) &\textit{0.027} (0.003) &\textbf{ 0.070} (0.004) & \textit{0.112} (0.005)\\
              & DeepHit &  0.053 (0.003) & 0.102 (0.007) & 0.141 (0.002)&  \textit{0.027} (0.003) & 0.072 (0.005) & 0.115 (0.005)\\
              & DeSurv& \textbf{0.049} (0.005) & \textbf{0.095} (0.009) & \textit{0.129} (0.003)&\textit{0.027} (0.003) & \textbf{0.070} (0.005) & 0.113 (0.004)\\
              & DSM & 0.057 (0.005) & 0.104 (0.006) & 0.141 (0.002) & \textit{0.027} (0.003) & \textit{0.071} (0.004) & \textbf{0.111} (0.004)\\
              & Fine-Gray & 0.057 (0.006) &  0.099 (0.007) &  0.131 (0.003) &\textit{0.027} (0.003) &  \textit{0.071} (0.005) &  \textit{0.112} (0.005) \\
              & CS Cox & 0.056 (0.006) &  \textit{0.098} (0.007) &  0.131 (0.003) &\textbf{0.026} (0.003) &  \textbf{0.070} (0.005) &  \textbf{0.111} (0.005)\\\midrule

            \parbox[t]{2mm}{\multirow{6}{*}{\rotatebox[origin=c]{90}{Synthetic}}} 
              & \textbf{NeuralFG} & \textbf{0.068} (0.003) &  \textit{0.125} (0.004) &  \textbf{0.192} (0.005) &\textbf{0.064} (0.003) & \textit{0.125} (0.002) & \textbf{0.191} (0.005)\\
              & DeepHit & 0.079 (0.003) & 0.136 (0.002) & \textit{0.212} (0.003) & 0.075 (0.003) & 0.132 (0.003) & \textit{0.204} (0.005) \\
              & DeSurv& \textbf{0.068} (0.002) &\textbf{0.124} (0.004) & \textbf{0.192} (0.004)& \textbf{0.064} (0.003) & \textbf{0.124} (0.003) & \textbf{0.191} (0.005)\\
              & DSM &  \textit{0.073} (0.002) & 0.139 (0.002) & \textit{0.220} (0.003) & \textit{0.069} (0.002) & 0.138 (0.002) & 0.217 (0.004)\\
              & Fine-Gray & 0.078 (0.002) &  0.159 (0.003) &  0.241 (0.002)& 0.074 (0.003) &  0.159 (0.003) &  0.238 (0.004)\\
              & CS Cox & 0.078 (0.002) &  0.159 (0.003) &  0.240 (0.002)& 0.074 (0.003) &  0.159 (0.003) &  0.238 (0.004) \\\midrule

            \parbox[t]{2mm}{\multirow{6}{*}{\rotatebox[origin=c]{90}{SEER}}} 
              & \textbf{NeuralFG} & \textbf{0.038} (0.000) & \textbf{0.069} (0.001) & \textbf{0.101} (0.000) &\textbf{0.009} (0.000) & \textit{0.021} (0.000) & \textbf{0.043} (0.000)\\
              & DeepHit &\textbf{0.038} (0.000) & \textit{0.070} (0.000) & \textit{0.102} (0.001) & \textbf{0.009} (0.000) & \textbf{0.020} (0.000) & \textbf{0.043} (0.000)\\
              & DeSurv& \textbf{0.038} (0.000) & \textit{0.070} (0.000) & \textit{0.102} (0.001) & \textbf{0.009} (0.000) & \textit{0.021} (0.000) & \textbf{0.043} (0.000)\\
              & DSM & \textit{0.039} (0.000) & 0.076 (0.001) & 0.112 (0.000) & \textbf{0.009} (0.000) & \textbf{0.020} (0.000) & \textbf{0.043} (0.000) \\
              & Fine-Gray &0.043 (0.001) &  0.081 (0.000) &  0.118 (0.000) & \textbf{0.009} (0.000) &  \textit{0.021} (0.000) &  \textit{0.044} (0.000) \\
              & CS Cox & 0.042 (0.001) &  0.081 (0.000) &  0.118 (0.000)&\textbf{0.009} (0.000) &  \textit{0.021} (0.000) &  \textit{0.044} (0.000)\\
        \end{tabular}}
    \caption{Comparison of the \textbf{Brier Score} across 5-fold cross-validation. Best performances are in \textbf{bold}, second best in \textit{italics}. \textit{NeuralFG is the model introduced in this paper.}}
    \label{tab:res:brier}
\end{table*}

\subsection{Cumulative evaluation}
\label{supp-app:cumulative}
The cumulative metrics summarise how a model performs over the total distribution. While having the advantage of representing performance in a single number, it is more disconnected from medical applications in which the risk horizon would be discretized to inform patients' treatment. Table~\ref{table:cumulative} displays the time-dependent C-index and cumulative time-dependent Brier score. These results echo the findings from Table~\ref{tab:result}.

\begin{table*}[p]
    \centering
        \centerline{\begin{tabular}{c|c|cc||cc}
            {}  &\multirow{2}{*}{Model} & \multicolumn{2}{c||}{Primary risk} & \multicolumn{2}{c}{Competing Risk} \\
             & &   C$^{td}$-Index  &      Brier Score&    C$^{td}$-Index  &    Brier Score  \\\midrule
            \parbox[t]{2mm}{\multirow{6}{*}{\rotatebox[origin=c]{90}{PBC}}} 
              & \textbf{NeuralFG} & 0.746 (0.116) & 0.166 (0.024)	 & \textit{0.785} (0.166) & \textit{0.154} (0.035) \\
              & DeepHit &  0.733 (0.069) & \textit{0.157} (0.013) 	 & 0.627 (0.088) & \textit{0.154} (0.013)  \\
              & DeSurv & \textit{0.804} (0.059) & \textit{0.157} (0.033)	 & \textbf{0.819} (0.123) & \textbf{0.153} (0.049)\\
              & DSM & \textbf{0.812} (0.050) & \textbf{0.152} (0.019) &0.707 (0.152) & 0.164 (0.028)  \\
              & Fine-Gray & 0.797 (0.057) & 0.182 (0.170) & 0.732 (0.138) & 0.177 (0.064) \\
              & CS Cox & 0.796 (0.056) & - & 0.769 (0.120) &0.160 (0.072) \\\midrule
              
            \parbox[t]{2mm}{\multirow{6}{*}{\rotatebox[origin=c]{90}{Framingham}}}        
              & \textbf{NeuralFG} & \textbf{0.775} (0.018) & \textit{0.089} (0.004) & \textit{0.716} (0.022) & 0.072 (0.002) \\
              & DeepHit & 0.760 (0.022) & 0.157 (0.141) & 0.698 (0.011) & 0.081 (0.003) \\
              & DeSurv &  \textit{0.771} (0.021) & \textbf{0.082} (0.041) & 0.712 (0.021) & 0.072 (0.003) \\
              & DSM & 0.767 (0.016) & 0.099 (0.002) & 0.701 (0.014) & \textbf{0.069} (0.002)  \\
              & Fine-Gray & 0.765 (0.016) & 0.152 (0.036)&\textit{0.716} (0.022)& 0.072 (0.003)	\\
              & CS Cox &0.767 (0.015) & - &\textbf{0.718} (0.028)&  \textit{0.071} (0.002)	\\\midrule

            \parbox[t]{2mm}{\multirow{6}{*}{\rotatebox[origin=c]{90}{Synthetic}}}
              & \textbf{NeuralFG} & \textbf{0.735} (0.010) &\textbf{0.228} (0.004)&\textbf{0.738} (0.014)&\textbf{0.233} (0.003)\\
              & DeepHit & 0.722 (0.009) &0.245 (0.004)&0.725 (0.010)& 0.240 (0.004)	 \\
              & DeSurv & \textit{0.734} (0.010) &\textit{0.231} (0.005)&\textit{0.737} (0.014) & \textit{0.237} (0.006)\\
              & DSM & 0.719 (0.010)&0.286 (0.005)&0.722 (0.017)& 0.287 (0.007) \\
              & Fine-Gray & 0.583 (0.007) &0.257 (0.002)&0.591 (0.014)&  0.265 (0.002)\\
              & CS Cox & 0.582 (0.007) &0.254 (0.002)&0.590 (0.013)& 0.262 (0.002)\\\midrule

            \parbox[t]{2mm}{\multirow{6}{*}{\rotatebox[origin=c]{90}{SEER}}}
              & \textbf{NeuralFG} & \textbf{0.819} (0.001) &\textbf{0.079} (0.000)&0.755 (0.004)& \textbf{0.032} (0.000) \\
              & DeepHit & 0.803 (0.002) &0.198 (0.004)&\textbf{0.763} (0.003)& \textit{0.148} (0.002)	\\
              & DeSurv & \textit{0.818} (0.001) &0.176 (0.002)&\textit{0.756} (0.004)&  0.183 (0.002)	\\
              & DSM & 0.801 (0.001)&0.193 (0.002)&0.745 (0.004)& 0.184 (0.001)  \\
              & Fine-Gray & 0.750 (0.002) &\textit{0.160} (0.033)&0.723 (0.004)& 0.179 (0.001)\\
              & CS Cox & 0.750 (0.002) &0.200 (0.003)&0.733 (0.005)&0.180 (0.001)\\
        \end{tabular}}
    \caption{Comparison of model performance by means (standard deviations) across 5-fold cross-validation. Best performances are in \textbf{bold}, second best in \textit{italics}. `-' indicates the divergence of the estimated Brier score. \textit{NeuralFG is the model introduced in this paper.}}
    \label{table:cumulative}
\end{table*}

\subsection{Implementation details}
The proposed experiments rely on the \verb|scikit-survival|~\citep{polsterl2020scikit}\footnote{\url{https://github.com/sebp/scikit-survival}} and \verb|pycox|\footnote{\url{https://github.com/havakv/pycox}} libraries for evaluation. For baselines' implementations, we used the R library \verb|riskRegression|\footnote{\url{https://github.com/tagteam/riskRegression}} for CS Cox and Fine-Gray, \verb|pycox| for DeepHit and \verb|auton-survival|~\citep{autonsurvival}\footnote{\url{https://github.com/autonlab/auton-survival}} for Deep Survival Machines.

\begin{table*}[!ht]
    \centering
        \centerline{\begin{tabular}{c|c|c|ccc||ccc}
            {} & \parbox[t]{2mm}{\multirow{2}{*}{\rotatebox[origin=c]{90}{Risk}}}  &\multirow{2}{*}{Model} & \multicolumn{3}{c||}{C-Index \textit{(Larger is better)}} & \multicolumn{3}{c}{Brier Score \textit{(Smaller is better)}} \\
             &  &  &           $q_{0.25}$ &           $q_{0.50}$ &           $q_{0.75}$ &  $q_{0.25}$ &           $q_{0.50}$ &           $q_{0.75}$ \\\midrule
             
            \parbox[t]{2mm}{\multirow{4}{*}{\rotatebox[origin=c]{90}{PBC}}} 
              & \parbox[t]{2mm}{\multirow{2}{*}{\rotatebox[origin=c]{90}{Dea.}}}
              & \textbf{NeuralFG} & 0.810 (0.079) &    0.795 (0.114) &    0.762 (0.123) &  0.099 (0.028) &    0.140 (0.020) &    0.169 (0.050) \\
              && \textit{MonoFG} &  \textbf{0.815} (0.086) &    \textbf{0.797} (0.097) &    \textbf{0.773} (0.114) &  \textbf{0.095} (0.026) &    \textbf{0.135} (0.026) &    \textbf{0.155} (0.060) \\\cdashline{3-9}
             
              & \parbox[t]{2mm}{\multirow{2}{*}{\rotatebox[origin=c]{90}{Tra.}}}
              & \textbf{NeuralFG} &  \textbf{0.799} (0.082) &    \textbf{0.709} (0.309) &    \textbf{0.788} (0.145) &\textbf{0.018} (0.001) &    \textbf{0.036} (0.015) &    \textbf{0.092} (0.017)  \\
              && \textit{MonoFG} & 0.699 (0.072) &    0.632 (0.272) &    0.709 (0.097) & \textbf{0.018} (0.001) &    0.040 (0.019) &    0.098 (0.019) \\\midrule
              
            \parbox[t]{2mm}{\multirow{4}{*}{\rotatebox[origin=c]{90}{Fram.}}} & \parbox[t]{2mm}{\multirow{2}{*}{\rotatebox[origin=c]{90}{CVD}}}
              & \textbf{NeuralFG} &  \textbf{0.872} (0.024) &  \textbf{0.812} (0.029) &  \textbf{0.782} (0.018) &  0.050 (0.003) &  \textbf{0.095} (0.010) &  \textbf{0.128} (0.004) \\
              && \textit{MonoFG} & 0.870 (0.024) &  0.807 (0.028) &  0.778 (0.020) &  \textbf{0.049} (0.003) &  \textbf{0.095} (0.009) &  \textbf{0.128} (0.005) \\\cdashline{3-9}

            & \parbox[t]{2mm}{\multirow{2}{*}{\rotatebox[origin=c]{90}{Dea.}}}
              & \textbf{NeuralFG} &  \textbf{0.745} (0.055) &  \textbf{0.717} (0.038) &  \textbf{0.713} (0.022) &  \textbf{0.027} (0.003) &  \textbf{0.070} (0.004) &  \textbf{0.112} (0.005)) \\
              && \textit{MonoFG} & 0.735 (0.047) &  \textbf{0.717} (0.037) &  \textbf{0.713} (0.018) &  \textbf{0.027} (0.003) &  0.071 (0.005) &  0.113 (0.005) \\\midrule

            \parbox[t]{2mm}{\multirow{4}{*}{\rotatebox[origin=c]{90}{Synthetic}}} & \parbox[t]{2mm}{\multirow{2}{*}{\rotatebox[origin=c]{90}{1}}}
              & \textbf{NeuralFG} & 0.791 (0.013) &  0.754 (0.013) &  \textbf{0.715} (0.011) &\textbf{0.068} (0.003) &  \textbf{0.125} (0.004) &  \textbf{0.192} (0.005)\\
              && \textit{MonoFG} & \textbf{0.792} (0.012) &  \textbf{0.755} (0.013) &  \textbf{0.715} (0.011) & \textbf{0.068} (0.003) &  \textbf{0.125} (0.004) &  \textbf{0.192} (0.006) \\\cdashline{3-9}
              
            & \parbox[t]{2mm}{\multirow{2}{*}{\rotatebox[origin=c]{90}{2}}}
              & \textbf{NeuralFG} & \textbf{0.801} (0.016) &  \textbf{0.755} (0.018) &  \textbf{0.714} (0.016) &\textbf{0.064} (0.003) &  \textbf{0.125} (0.002) &  \textbf{0.191} (0.005) \\
              && \textit{MonoFG} & \textbf{0.801} (0.015) &  \textbf{0.755} (0.016) &  0.713 (0.013) & \textbf{0.064} (0.003) &  \textbf{0.125} (0.002) &  \textbf{0.191} (0.004) \\\midrule

            \parbox[t]{2mm}{\multirow{4}{*}{\rotatebox[origin=c]{90}{SEER}}} & \parbox[t]{2mm}{\multirow{2}{*}{\rotatebox[origin=c]{90}{BC}}}
              & \textbf{NeuralFG} & 0.893 (0.002) &  \textbf{0.855} (0.001) &  \textbf{0.815} (0.001) & \textbf{0.038} (0.000) &  \textbf{0.069} (0.001) &  \textbf{0.101} (0.000) \\
              && \textit{MonoFG} &  \textbf{0.894} (0.001) &  \textbf{0.855} (0.001) &  \textbf{0.815} (0.001) & \textbf{0.038} (0.000) &  \textbf{0.069} (0.000) &  \textbf{0.101} (0.001) \\\cdashline{3-9}

            & \parbox[t]{2mm}{\multirow{2}{*}{\rotatebox[origin=c]{90}{CVD}}}
              & \textbf{NeuralFG} &  0.799 (0.010) &  0.782 (0.005) &  \textbf{0.758} (0.003)& \textbf{0.009} (0.000) & \textbf{0.021} (0.000) & \textbf{0.043} (0.000) \\
              && \textit{MonoFG} &  \textbf{0.804} (0.010) &  \textbf{0.785} (0.005) &  \textbf{0.758} (0.004) & \textbf{0.009} (0.000) & \textbf{0.021} (0.000) & \textbf{0.043} (0.000) \\
        \end{tabular}}
    \caption{Comparison of model performance by means (standard deviations) across 5-fold cross-validation. Best performances are in \textbf{bold}.}
    \label{supp-tab:app:mono}
\end{table*}

\section{Using $R$ outcomes vs. $R$ networks}
\label{supp-app:mono}
In this section, we investigate the impact of using multiple networks -- one for each competing risk -- instead of one network with multiple outcomes. The model \textbf{MonoFG} consists of the same architecture presented in Figure~\ref{fig:nfg} with only one monotonic network with $R$ outputs. Table~\ref{supp-tab:app:mono} shows limited differences between the two architectures. However, we encourage the use of multiple networks when the competing risks present large distributional differences.

\begin{table*}[!ht]
    \centering
         \centerline{\begin{tabular}{c|c|c|ccc||ccc}
            {} & \parbox[t]{2mm}{\multirow{2}{*}{\rotatebox[origin=c]{90}{Risk}}}  &\multirow{2}{*}{Model} & \multicolumn{3}{c||}{C-Index \textit{(Larger is better)}} & \multicolumn{3}{c}{Brier Score \textit{(Smaller is better)}} \\
             &  &  &           $q_{0.25}$ &           $q_{0.50}$ &           $q_{0.75}$ &  $q_{0.25}$ &           $q_{0.50}$ &           $q_{0.75}$ \\\midrule
             
            \parbox[t]{2mm}{\multirow{8}{*}{\rotatebox[origin=c]{90}{Synthetic}}} 
              & \parbox[t]{2mm}{\multirow{4}{*}{\rotatebox[origin=c]{90}{1}}}
              & {$n = 1$} &  0.779 (0.012) &  0.743 (0.013) &  0.705 (0.010) & \textbf{0.076} (0.003) &  \textbf{0.180} (0.004) &  \textbf{0.344} (0.004) \\
              && $\mathbf{n = 15}$ &  \textbf{0.792} (0.011) &  \textbf{0.758} (0.014) &  \textbf{0.724} (0.012) & 0.079 (0.003) &  0.186 (0.004) &  0.355 (0.004) \\
              && {$n = 100$} &  0.791 (0.013) &  \textbf{0.758} (0.014) &  0.723 (0.012) & 0.079 (0.003) &  0.186 (0.004) &  0.354 (0.004) \\
              && $n=1,000$ & \textbf{0.792} (0.011) &  \textbf{0.758} (0.013) &  0.723 (0.011) & 0.079 (0.003) &  0.186 (0.004) &  0.355 (0.004) \\
              \cdashline{3-9}
             & \parbox[t]{2mm}{\multirow{4}{*}{\rotatebox[origin=c]{90}{2}}}
              & {$n = 1$} &  0.788 (0.016) &  0.737 (0.021) &  0.702 (0.017) & \textbf{0.073} (0.003) &  \textbf{0.180} (0.005) &  \textbf{0.338} (0.009) \\
              && $\mathbf{n = 15}$ &  0.800 (0.014) &  \textbf{0.754} (0.017) &  \textbf{0.721} (0.016) & 0.074 (0.003) &  0.185 (0.005) &  0.346 (0.009) \\
              && {$n = 100$} &  0.800 (0.013) &  0.753 (0.016) &  0.720 (0.015) & 0.074 (0.003) &  0.185 (0.004) &  0.346 (0.008) \\
              && $n=1,000$ & \textbf{0.801} (0.014) &  0.753 (0.017) &  0.720 (0.017) & 0.075 (0.003) &  0.185 (0.004) &  0.347 (0.008) \\
        \end{tabular}}
    \caption{Impact of increasing $n$ on DeSurv performances. Performance measured by means (standard deviations) across 5-fold cross-validation. Best performances are in \textbf{bold}.}
    \label{supp-tab:app:n}
\end{table*}

\begin{table*}[!ht]
\centering
    \begin{tabular}{c|ccc}
                       & Convergence Speed         & Total Training Time\\
                      &(in number of iterations)   & (in seconds) \\\midrule
    \textbf{NeuralFG}&  91.98 (43.33) & 13.60 (6.03)  \\
    \textit{MonoFG}&  66.26 (28.08) & \textbf{6.66} (2.90)  \\\cdashline{2-3}
    DeSurv ($n = 1$)   &  151.88 (123.50) & 13.93 (11.07) \\
    DeSurv ($\mathbf{n = 15}$) & 55.09 (43.56)& 56.68 (47.35) \\
    DeSurv ($n = 100$) & \textbf{52.02} (24.45)&  363.95 (172.55)
    \end{tabular}
    \caption{Training speed comparison on the Framingham dataset. Performance measured by means (standard deviations) across 100-fold \textit{Monte Carlo} cross-validation.}
    \label{supp-tab:app:speed}
\end{table*}

\section{DeSurv}
\subsection{Impact of $n$}
\label{supp-app:comparison}
In the upper limit, the Gauss-Legendre quadrature would lead to the exact estimation of the likelihood. However, this requires $n$ forward passes in the neural network with $n$, the number of point estimation. Fixing the architecture to a 3 hidden layer perceptron with 50 nodes, we measure the model's performances for $n$ in $[1, 15, 100, 1000]$ on the Synthetic dataset as shown in Table~\ref{supp-tab:app:n}. For $n = 1$, NeuralFG and DeSurv present the same computational complexity. However, DeSurv benefits from larger $n$. Note that there is limited gain above the recommended 15-degree quadrature.

\subsection{Training speed}
\label{supp-app:speed}
Finally, we examine the training and convergence speed for both DeSurv and NeuralFG on the Framingham dataset. We trained a fixed architecture with a total depth of 3 hidden layers composed of 50 nodes each. The learning rate was fixed at $0.001$ and the batch size at $100$. Table~\ref{supp-tab:app:speed} presents the number of training iterations required to converge and the training time over 100 random splits of the data. We parallelised DeSurv's $n$ forward passes following the original paper's recommendation. This set of experiments is performed on an Apple M1 Pro chip with 32 GB of memory.

The Desurv's results highlight that a coarser approximation ($n = 1$) requires more iterations to converge due to the lower-quality target loss, but each iteration is faster. Conversely, increasing $n$ results in fewer iterations for convergence, but slower training. Echoing the theoretical computational cost introduced in Section~\ref{sec:computationalcost}, our proposed methodology results in faster iterations, especially when considering a single network architecture for competing risks as shown by MonoFG's training time. However, the larger number of iterations required by our proposed methods in comparison to DeSurv reflects the more complex convergence of \emph{constrained} monotonic neural networks.

\end{document}